\documentclass[10pt,a4paper,fleqn]{article}

\usepackage[
    top=20mm,
    bottom=20mm,
    left=18mm,
    right=18mm
]{geometry}

\usepackage{graphicx}
\usepackage{multicol}
\usepackage{titlesec}
\usepackage{fancyhdr}
\usepackage{setspace}
\usepackage{array}
\usepackage{booktabs}
\usepackage{tikz}
\usepackage{tcolorbox}
\usepackage{amssymb}
\usepackage{caption}
\usepackage{fontspec}
\usepackage{amsmath}
\usepackage{graphicx}
\usepackage{siunitx}
\usepackage[numbers]{natbib}

\usepackage{hyperref}
\usepackage{url}
\usepackage{lipsum}

\usepackage{booktabs}
\usepackage{titlesec}
\titleformat{\section}
    {\normalsize\bfseries}
    {\thesection}
    {1em}
    {}

\titleformat{\subsection}
  {\footnotesize\bfseries}
  {\thesubsection}
  {1em}
  {}

\titleformat{\subsubsection}
  {\footnotesize\bfseries}
  {\thesubsubsection}
  {1em}
  {}

\titleformat{\paragraph}[runin]
  {\footnotesize\bfseries}
  {\theparagraph}
  {1em}
  {}

\usepackage{lipsum}

\usepackage{varwidth}
\usepackage{adjustbox}

\definecolor{tether-gray}{HTML}{F2F1EF}

\newtcolorbox[auto counter]{custombox}[2][]{
    title={Box~\thetcbcounter: #2},
    label={#1},
    fonttitle=\bfseries,
    coltitle=black,
    before title=\vspace{15pt},
    colback=tether-gray,
    colframe=tether-gray,
    boxrule=0pt,
    arc=5pt,
    left=15pt,
    right=15pt,
    bottom=15pt,
    width=\textwidth,    
}

\newcommand{\tablebox}[1]{
    \tcbox[
        colback=tether-gray,
        colframe=tether-gray,
        boxrule=0pt,
        left=15pt,
        right=15pt,
        top=15pt,
        bottom=15pt
    ]{#1}
}

\begin{document}


\begin{flushleft}
\includegraphics[height=20pt]{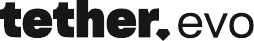}
\end{flushleft}

\vspace{10pt}


\begin{tcolorbox}[
    colback=tether-gray,
    colframe=tether-gray,
    boxrule=0pt,
    arc=5pt,
    left=15pt,
    right=15pt,
    top=15pt,
    bottom=15pt,
    width=\textwidth
]


{\huge\bfseries\begin{flushleft}
Decoding the decoder: Contextual sequence-to-sequence modeling for intracortical speech decoding\end{flushleft}}



{\small\bfseries
Michal Olak\textsuperscript{1},
Tommaso Boccato\textsuperscript{1},
Matteo Ferrante\textsuperscript{1}
}

\vspace{2.5pt}

{\small
\textsuperscript{1}Tether Evo
}

\vspace{15pt}


{\footnotesize
\textbf{Objective.} Speech brain--computer interfaces require decoders that translate intracortical attempted-speech activity into linguistic output while remaining robust to limited data and day-to-day variability. While prior high-performing systems have largely relied on framewise phoneme decoding combined with downstream language models, it remains unclear what contextual sequence-to-sequence decoding contributes to sublexical neural readout, robustness, and interpretability.
\textbf{Approach.} We evaluated a multitask Transformer-based sequence-to-sequence model for attempted speech decoding from area 6v intracortical recordings. The model jointly predicts phoneme sequences, word sequences, and auxiliary acoustic features. To address day-to-day nonstationarity, we introduced the Neural Hammer \& Scalpel (NHS) calibration module, which combines global alignment with feature-wise modulation. We further analyzed held-out-day generalization and attention patterns in the encoder and decoders.
\textbf{Main results.} On the  Willett et al. \cite{Willett2023Dataset} dataset, the proposed model achieved a state-of-the-art phoneme error rate of 14.3\%. Word decoding reached 25.6\% WER with direct decoding and 19.4\% WER with candidate generation and rescoring. NHS substantially improved both phoneme and word decoding relative to linear or no day-specific transform, while held-out-day experiments showed increasing degradation on unseen days with temporal distance. Attention visualizations revealed recurring temporal chunking in encoder representations and distinct use of these segments by phoneme and word decoders.
\textbf{Significance.} These results indicate that contextual sequence-to-sequence modeling can improve the fidelity of neural-to-phoneme readout from intracortical speech signals, even when word-level performance remains competitive rather than state-leading. They further identify session nonstationarity as a major barrier for robust speech BCIs and suggest that attention-based analyses can generate useful hypotheses about how neural speech evidence is segmented and accumulated over time.
}

\end{tcolorbox}

\vspace{10pt}


\footnotesize


\begin{figure}[h]
    \centering
    \includegraphics[width=1.0\linewidth]{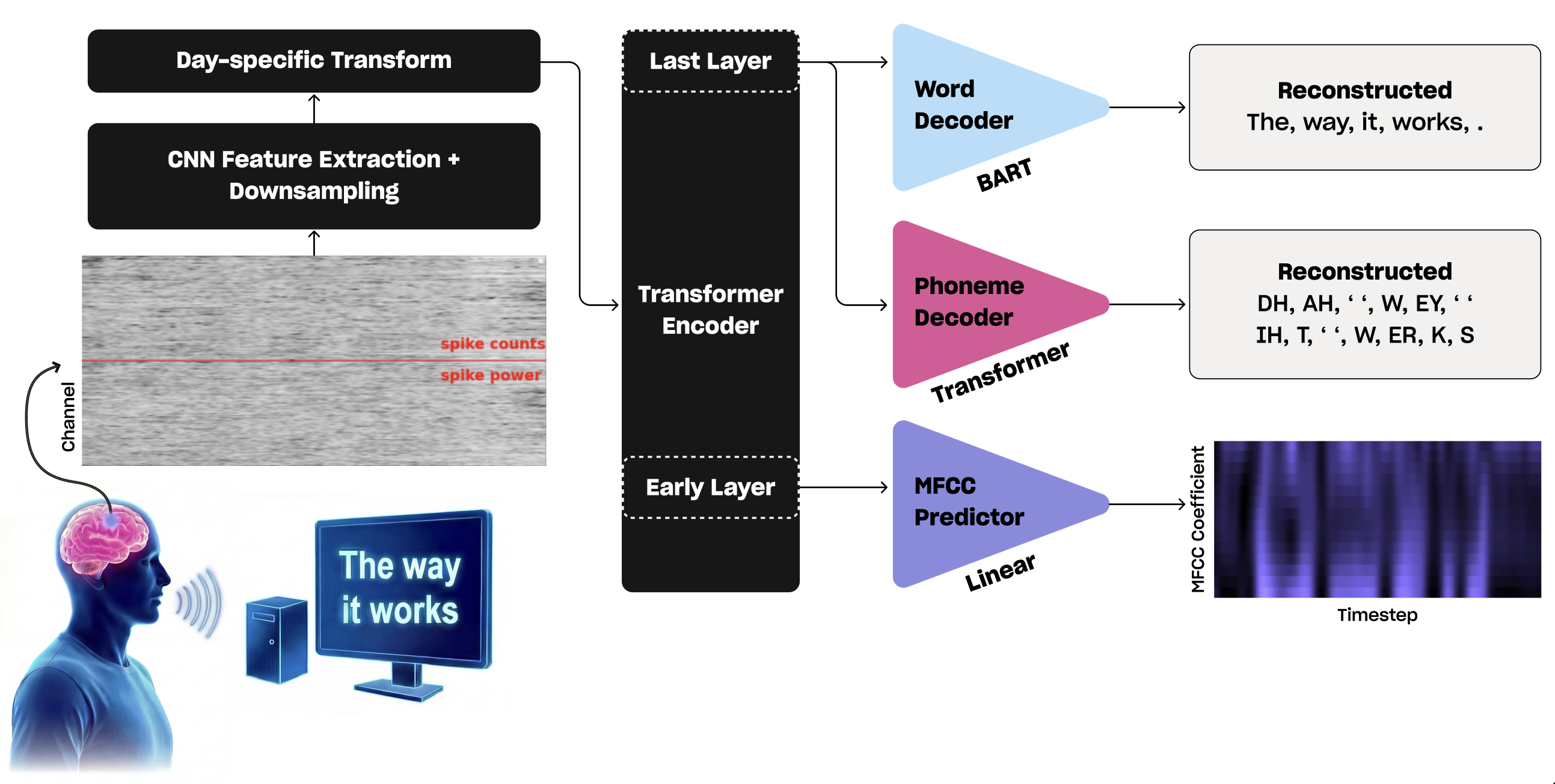}
    \caption{\textbf{Model overview.} Spike counts and spike-band power are processed in parallel convolutional branches, merged into a shared latent sequence, and reweighted by a lightweight content-dependent gate. The \emph{NHS} day-specific transform then aligns the sequence by combining a global affine transformation (\emph{hammer}) with FiLM-style feature-wise modulation (\emph{scalpel}) through a learned gate. The calibrated representation is encoded by a 6-layer Transformer and decoded by three heads: an autoregressive phoneme decoder, a BART-based word decoder, and an auxiliary MFCC prediction head attached to an intermediate encoder layer.}
    \label{fig:scheme}
\end{figure}

\section{Introduction}

Decoding speech from neural recordings offers a pathway to restore communication for individuals with severe neurological and neuromuscular conditions, such as amyotrophic lateral sclerosis (ALS) or locked-in syndrome. Loss of speech drastically reduces quality of life, making speech-oriented brain--computer interfaces (BCIs) an important clinical goal. Recent advances have clarified how articulatory and linguistic information is represented across speech-related cortex ~\citep{Bouchard2013,Chen2024Flinker,Goldstein2025} and have culminated in initial demonstrations of real-time speech neuroprostheses for communication restoration~\citep{Moses2021,Willett2023,Card2024}. However, translating these proof-of-concept systems into robust clinical devices remains limited by the scarcity of standardized human neural datasets, the computational complexity of decoding pipelines, and severe day-to-day nonstationarity in chronic intracortical microelectrode array recordings.

Current high-performing speech BCI decoders predominantly use recurrent neural networks (RNNs) trained with a connectionist temporal classification (CTC) objective~\citep{Moses2021,Willett2023,Card2024}. In this paradigm, phoneme probabilities are predicted framewise under a conditional-independence assumption, often from narrow temporal windows, and long-range structure is recovered mainly through downstream language modeling. Recent work has further reduced word error rate (WER) by combining context-aware intermediate representations with large language models~\citep{Li2025}. While these approaches have produced impressive communication-level performance, they also sharpen an important distinction: WER reflects both the neural decoder and the linguistic prior applied after it. By contrast, phoneme error rate (PER) more directly measures how faithfully neural activity is mapped onto speech-relevant sublexical structure. This distinction is especially relevant for speech motor cortical recordings, where neural activity is expected to primarily reflect articulatory and motor planning variables, rather than high-level semantic content~\citep{Bouchard2013}.

Sequence-to-sequence (seq2seq) modeling offers an alternative to framewise CTC decoding. In a seq2seq \emph{encoder--decoder} architecture, an encoder first converts the entire neural time series into a contextual representation, and a decoder then generates output tokens autoregressively, conditioning each prediction on both the encoded input and the previously generated outputs. This removes the conditional-independence and strict locality assumptions of CTC and allows the model to learn more flexible alignments between neural evidence and linguistic units. Transformer-based seq2seq models implement this using attention: \emph{self-attention} lets the encoder integrate information across time by allowing each timestep to attend to other timesteps in the input, while \emph{cross-attention} lets the decoder selectively retrieve the most relevant portions of the encoded neural sequence when producing each token~\citep{vaswani2017attention}. This encoder--decoder formulation is prominent in machine translation, where a source sequence is mapped to a target sequence; here, we similarly ``translate'' neural activity into a sequence of phonemes or words. Beyond performance, attention provides a qualitative window into how neural information is segmented and aggregated during decoding, allowing us to ``decode the decoder.'' Rather than treating the decoder purely as a black box, this enables analysis of emergent decoding patterns and may generate hypotheses about the temporal organization of neural speech representations.

In this work, we use a Transformer-based seq2seq model to test whether fully contextual decoding improves intracortical speech readout relative to CTC-based decoders. This added flexibility comes with a challenge: attention-based models must learn more of the task structure from limited data, and neural representations can drift substantially across recording days. To mitigate these issues, we adopt a multitask training setup that jointly predicts phoneme and word sequences, with auxiliary acoustic targets (MFCCs) to shape intermediate representations. We additionally introduce the Neural Hammer \& Scalpel (NHS) day-specific calibration module to address cross-day nonstationarity. Evaluated on the Willett et al.\cite{Willett2023Dataset} dataset, our model achieves a state-of-the-art PER of 14.3\% and reveals recurring attention patterns, including emergent temporal chunking and differentiated use of neural context by phoneme and word decoders. Finally, we provide a scaling analysis to estimate how performance changes with dataset size and to contextualize data requirements for future improvements in intracortical speech decoding. Together, these results position seq2seq models not only as an alternative decoding framework, but also as useful tools for studying how speech-relevant neural information is organized over time.

\section{Methods}\label{sec:methods}


\subsection{Dataset and preprocessing}\label{sec:dataset}

We evaluated all models on the intracortical speech BCI dataset released by Willett et al.\cite{Willett2023Dataset}. The dataset contains neural recordings from a participant with amyotrophic lateral sclerosis, implanted with four 64-channel Utah arrays in the speech motor cortex. To keep the modeling setting comparable to established baselines, we restricted our analyses to the 128 channels in area 6v (ventral premotor cortex), consistent with prior evidence that this region carries the strongest speech-decoding signal. Over 24 recording days, the participant attempted to produce 12{,}100 prompted sentences drawn from a large-vocabulary corpus. Each trial consisted of an instructed delay period (planning) followed by a go cue to initiate attempted speech; we used only neural activity after the go cue. The dataset provides two neural features per channel in 20\,ms bins: multi-unit threshold crossings (spike counts) and high-frequency spike-band power ($>250$\,Hz). Each feature was z-scored within the recording block to reduce session-specific offsets. We utilized the standard train and test partitions, omitting the competition holdout.

Targets were extracted at three levels: (1) 39-symbol ARPAbet phoneme sequences, (2) word-level targets tokenized via the BART tokenizer~\citep{wolf2020transformers}, and (3) 14-dimensional Mel-frequency cepstral coefficients (MFCCs). Because the time-synchronized microphone audio is unintelligible but still preserves coarse onset and envelope information, MFCCs were utilized as auxiliary acoustic supervision to regularize intermediate neural representations.

\subsection{Model overview}\label{sec:model}

Our model is a multitask sequence-to-sequence architecture that maps intracortical activity to phoneme and word sequences while also predicting auxiliary acoustic features. It consists of four components: (1) a convolutional front end for feature extraction and temporal downsampling, (2) a day-specific calibration module, (3) a Transformer encoder, and (4) task-specific output heads.

Let the neural input be $X_0 \in \mathbb{R}^{T \times C}$, where $C=256$ corresponds to 128 channels with two features per channel. Spike counts and spike-band power are first processed in separate one-dimensional convolutional branches and merged into a shared latent sequence. Temporal downsampling is performed by stride-4 convolutions, producing a representation at an effective 80\,ms cadence. We denote the resulting latent sequence by
\[
X_{\mathrm{feat}} \in \mathbb{R}^{L \times D},
\]
where $L \approx T/4$ and $D=512$. A lightweight content-dependent gate reweights latent channels frame by frame before calibration. 

The calibrated representation is then passed through a 6-layer Transformer encoder ($D_{\text{model}}=512$, 8 attention heads, feedforward dimension 2048). Three output heads are attached. The phoneme head is an autoregressive Transformer decoder that predicts phoneme tokens. The word head uses the decoder portion of \texttt{bart-base}~\citep{lewis2020bart}, with the first three layers frozen and the remaining layers fine-tuned for neural conditioning. The auxiliary MFCC head is a linear projection from an intermediate encoder layer and is used only during the first training stage.

\subsection{Day-specific calibration module}\label{sec:nhs}

To address day-to-day nonstationarity, we introduced a day-specific calibration module termed Neural Hammer \& Scalpel (NHS). The module operates on the convolutional feature sequence and assumes that the recording day identity is known at inference time. NHS combines two complementary day-conditioned transformations: a global affine alignment that can remix latent features across dimensions, and a FiLM-style feature-wise modulation that provides milder per-feature adjustment.

Let $X \in \mathbb{R}^{L \times D}$ denote the front-end output for one trial and let $d$ index the recording day. The hammer branch applies a day-specific affine transformation,
\[
X_{\mathrm{h}} = XW_d + \mathbf{1}b_d^\top,
\]
where $W_d \in \mathbb{R}^{D \times D}$ and $b_d \in \mathbb{R}^{D}$ are learned day-specific parameters. The scalpel branch applies a day-conditioned feature-wise modulation,
\[
X_{\mathrm{s}} = X \odot \gamma_d + \beta_d,
\]
where $\gamma_d,\beta_d \in \mathbb{R}^{D}$ are produced from a learned day embedding through a small multilayer perceptron. The two branches are then combined by a learned gate,
\[
\hat{X} = \phi\!\left(g_d X_{\mathrm{h}} + (1-g_d)X_{\mathrm{s}}\right), \qquad g_d \in (0,1),
\]
where $\phi(\cdot)$ is a smooth nonlinearity. The motivation for this design is that chronic intracortical recordings exhibit both broad cross-session drift and more localized feature-specific variability. The global affine branch is intended to correct coarse shifts in the latent space, whereas the feature-wise branch captures lighter-weight recalibration. In ablation experiments, we compared NHS against a simpler linear day transform matching prior work and against models without any day-specific transform. 

\subsection{Baseline model}\label{sec:baseline}

To isolate the effect of the sequence-to-sequence formulation, we implemented a controlled recurrent baseline. This model used the same convolutional front end, optional MFCC supervision, and day-specific transform as the main model, but replaced the Transformer encoder and autoregressive phoneme decoder with a gated recurrent unit (GRU) encoder trained with a CTC objective for phoneme prediction. Word-level decoding was performed using the same BART-based head for comparability. This baseline was intended as a controlled architectural comparison rather than an exact reimplementation of previously published systems.

\subsection{Word-level rescoring}\label{sec:rescoring}
In addition to direct word decoding, we evaluated an optional post hoc rescoring pipeline designed to improve word-level output selection. For each sentence, the decoder generated a set of candidate hypotheses. These candidates were rescored using three signals: (1) consistency with the model's phoneme prediction, quantified through phoneme error rate after grapheme-to-phoneme conversion of each candidate; (2) the autoregressive phoneme decoder likelihood of the candidate phoneme sequence conditioned on the neural input; and (3) an external language-model score reflecting linguistic plausibility. A linear blend of these scores, with coefficients tuned on held-out validation data, selected the final prediction for each trial. This rescoring stage was used only for word-level evaluation and did not alter the underlying phoneme decoder.

\subsection{Training procedure}\label{sec:training}

Models were trained in up to two stages. For phoneme-only and phoneme+MFCC variants, the encoder, phoneme decoder, and optional MFCC head were trained jointly. For models that included word decoding, Stage~1 optimized the encoder together with the phoneme and MFCC heads, and Stage~2 replaced the MFCC head with the BART word decoder and continued joint training of phoneme and word objectives. This staged procedure was intended to first stabilize sublexical and acoustic structure before introducing the more linguistically informed word-level objective. Optimization used AdamW with learning-rate reduction on plateau based on validation phoneme error rate. Regularization included dropout, time masking, and channel masking. Model selection was based on validation PER. The same train/validation/test partitions were used across ablations and baseline comparisons.

\subsection{Evaluation metrics and reporting}\label{sec:metrics}

We report phoneme error rate (PER) and word error rate (WER), both computed as normalized Levenshtein distance. PER was computed after removing sequence boundary tokens. WER was computed after lowercasing and removing punctuation. Direct decoding and rescored decoding were reported separately for word-level results.

To assess robustness to session variability, we also evaluated held-out-day generalization by applying day transforms from temporally neighboring sessions to unseen days and measuring the resulting performance degradation. Unless otherwise stated, results were aggregated over repeated runs with different random seeds, with mean and standard deviation reported in the Results section.

\section{Results}\label{sec:results}

\begin{table}[h]
\centering
\footnotesize
\caption{Comparison of phoneme- and word-level decoding performance. Published reference systems are shown separately from controlled internal baselines and day-specific transform ablations. Our results are reported as mean $\pm$ SD across 5 random seeds. Published results are taken from the respective papers and are not directly seed-matched to our runs.}
\label{tab:decoding_results}
\tablebox{
\begin{tabular}{lcc}
\toprule
\textbf{Model} & \textbf{PER (\%)} & \textbf{WER (\%)} \\
\midrule
Willett et al.\ (2023) RNN--CTC + LM & 19.7 & 17.4 \\
Li et al. (2025) DCoND-LIFT & 15.34 & \textbf{5.77} \\
\midrule
RNN--CTC (ours) & 17.4 $\pm$ 0.8 & -- \\
\quad + MFCC & 17.0 $\pm$ 0.3 & -- \\
\quad + BART & 21.0 $\pm$ 0.7 & 30.9 $\pm$ 0.8 \\
\quad + MFCC + BART & 17.0 $\pm$ 0.5 & 29.0 $\pm$ 0.4 \\
Transformer Seq2Seq & 14.8 $\pm$ 0.3 & -- \\
\quad + MFCC & 14.6 $\pm$ 0.1 & -- \\
\quad + BART & 14.5 $\pm$ 0.3 & 26.0 $\pm$ 0.4 \\
\quad + MFCC + BART & \textbf{14.3 $\pm$ 0.3} & 25.6 $\pm$ 0.2 \\
\quad + MFCC + BART + rescoring & 14.3 $\pm$ 0.3 & 19.4 $\pm$ 0.3 \\
\midrule
Transformer Seq2Seq + BART + Linear DT & 17.2 $\pm$ 0.5 & 28.6 $\pm$ 0.4 \\
Transformer Seq2Seq + BART + No DT & 18.1 $\pm$ 0.6 & 30.4 $\pm$ 0.5 \\
\bottomrule
\end{tabular}
}
\end{table}




Table~\ref{tab:decoding_results} summarizes decoding performance across published reference systems, controlled internal baselines, and day-transform ablations. The strongest result of our model is at the phoneme level. The best multitask seq2seq variant (Transformer + MFCC + BART) achieved a phoneme error rate of 14.3 $\pm$ 0.3\%, improving over our controlled RNN--CTC baseline (17.4 $\pm$ 0.8\%), the previously reported Willett et al.\ hybrid baseline (19.7\%), and the recent context-aware diphoneme system of Li et al.\ (15.34\%). These results indicate that fully contextual sequence-to-sequence decoding improves sublexical neural readout beyond both monophone-based framewise CTC decoding and diphoneme-enriched CTC pipelines.

At the word level, the picture is different. Our direct seq2seq model reached 25.6 $\pm$ 0.2\% WER, which improved to 19.4 $\pm$ 0.3\% when candidate generation and rescoring were added. This substantially narrows the gap to the Willett et al.\ two-stage hybrid pipeline (17.4\% WER), but remains above the strongest published multi-stage systems, including Li et al.'s DCoND-LIFT result (5.77\%), which relies on heavier multistage language-model-assisted refinement and ensembling. Notably, our direct decoder operates end-to-end without WFST-style search or other heavy offline decoding stacks, yielding a relatively lightweight pipeline that reduces the gap to real-time deployment constraints. Overall, we interpret our main performance contribution as improved phoneme-level fidelity rather than state-leading final text accuracy.

Within our controlled ablations, multitask supervision improved performance consistently. Adding MFCC supervision lowered PER in both the recurrent and seq2seq model families, while adding the word-level decoder further improved phoneme decoding and enabled direct word generation. The full Transformer + MFCC + BART configuration provided the best performance without rescoring. Additional ablations of BART initialization and training are reported in Appx.~\ref{bart init}; pretrained and randomly initialized BART decoders yielded similar performance, suggesting that much of the useful structure in this regime is learned from the supervised neural-decoding task itself.


\subsection{Day-specific calibration and held-out-day generalization}\label{subsec:generalization}

Replacing the proposed \emph{Neural Hammer \& Scalpel} (NHS) transform with either the global linear day transform introduced by Willet et al.\cite{Willett2023} or no day-specific transform degraded both PER and WER (Table~\ref{tab:decoding_results}). Relative to the full model, the linear transform increased error by 2.7 percentage points in PER and 2.6 percentage points in WER, while removing day-specific calibration altogether increased error by 3.6 PER points and 4.4 WER points. These ablations indicate that flexible session-specific calibration is an important contributor to both sublexical and word-level decoding.

\begin{figure}[h]
    \centering
    \includegraphics[width=0.9\linewidth]{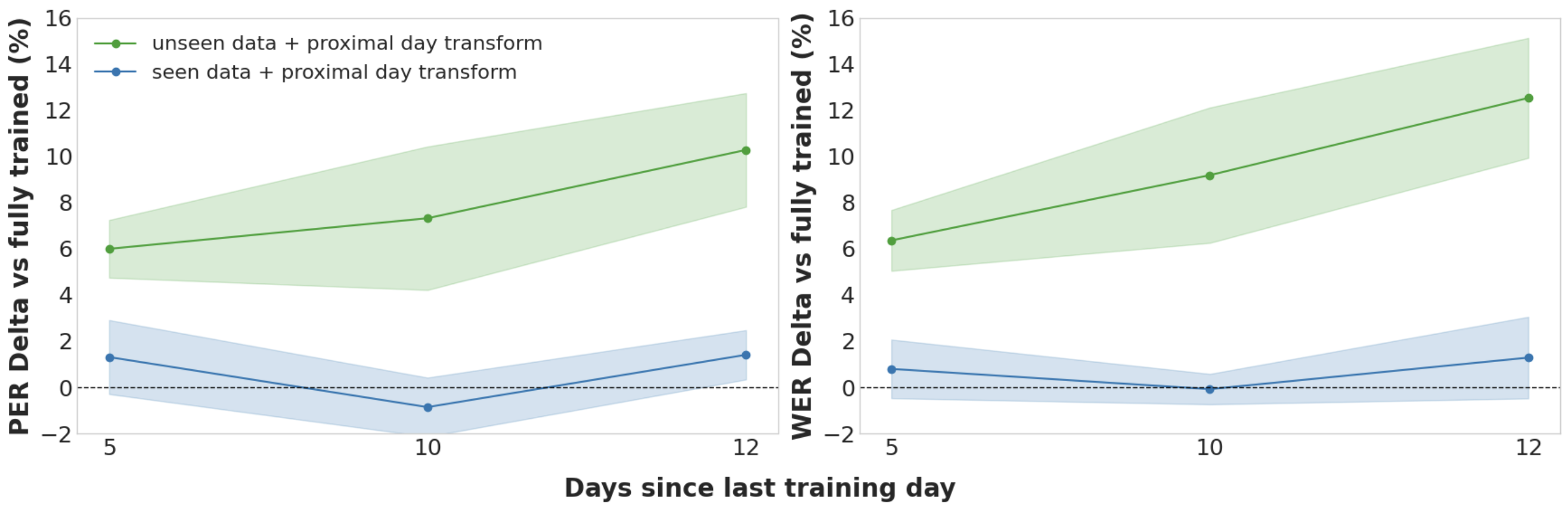}
    \caption{\textbf{Held-out-day generalization.} PER/WER deltas on held-out days 22--24 relative to a 24-day model evaluated with day-specific transforms (0\%). Blue curves (\emph{seen data + proximal}) show the same model re-evaluated on these days using the transform from day~21. Green curves (\emph{unseen data + proximal}) show a 21-day model evaluated on days 22--24 with the day-21 transform, after subtracting the mean 21- vs.\ 24-day performance gap on days 1--21. Mean over 5 seeds, shaded regions denote 95\% bootstrap confidence intervals.}
    \label{fig:generalization}
\end{figure}

To assess robustness to out-of-distribution neural data, we evaluated held-out-day generalization using models trained either on all 24 recording days or on the first 21 days only. As a reference, the 24-day model was first evaluated on days 22--24 using the correct day-specific transforms. We then reused the transform from the last training day (day 21) in two conditions: (i) the 24-day model re-evaluated on days 22--24 (\emph{seen data + proximal}), and (ii) the 21-day model evaluated on days 22--24 with the day-21 transform (\emph{unseen data + proximal}).

On days included in training, substituting the proximal transform changed PER and WER by at most approximately 2 percentage points, indicating that neighboring-day NHS transforms are often similar. For unseen days, however, the performance penalty started at roughly 6 percentage points and increased to approximately 10--13 percentage points as the temporal gap grew. These results show that day-specific calibration substantially improves decoding, but does not fully remove across-day distribution shift. In its current form, NHS reduces rather than eliminates the calibration burden, leaving meaningful residual drift for the shared encoder--decoder to handle.

\subsection{Attention structure and emergent temporal organization}\label{subsec:attention}

\begin{figure}[h]
    \centering
    \includegraphics[width=1.0\linewidth]{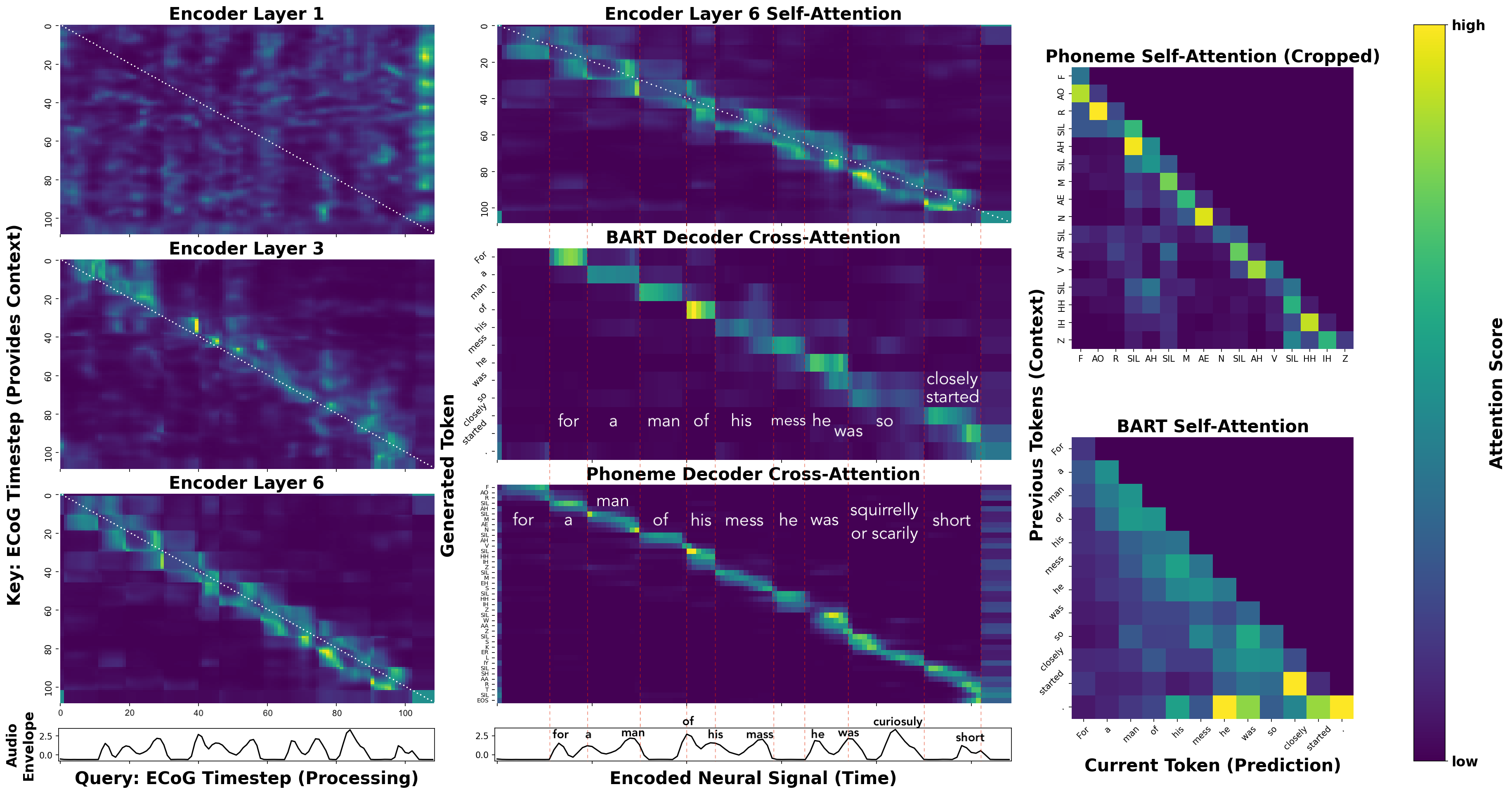}
    \caption{\textbf{Attention analysis.} Left: encoder self-attention for layers 1, 3, and 6. Middle: cross-attention for the BART and phoneme decoders. Right: decoder self-attention. Dotted vertical lines align with salient envelope events. Late encoder layers exhibit clean diagonals and ``boxed'' segments that decoders repeatedly attend to; phoneme self-attention is local, while BART self-attention is broadly triangular. The trial text is \emph{``for a man of his mass he was curiously short''}; target words are approximately aligned to envelope features, as no ground-truth timing annotations are available in this dataset.}
    \label{fig:attention}
\end{figure}

Because the dataset does not provide explicit temporal alignments between neural activity and phoneme or word units, the seq2seq model must learn its own internal organization of time. We therefore analyzed encoder self-attention, decoder cross-attention, and decoder self-attention to examine how neural evidence is segmented, retrieved, and accumulated during decoding. Figure~\ref{fig:attention} shows a representative trial from the validation set.

In the encoder, early layers distribute attention broadly, whereas later layers develop clearer local structure. By layer 3, diagonal patterns and compact local clusters begin to emerge. By layer 6, these patterns often take the form of short boxed segments separated by sharper transitions, suggesting that the encoder organizes the trial into temporally localized chunks. These boundaries qualitatively track salient events in the neural envelope, such as onsets and offsets, although we emphasize that the dataset lacks precise speech-unit timing annotations.

The two decoders re-use these encoder-derived segments in different ways. The phoneme decoder tends to focus locally within a segment and often progresses through a short temporal cascade, consistent with finer-grained sublexical resolution. The word decoder usually concentrates its evidence more coarsely, often over one dominant segment or a small number of adjacent segments, consistent with broader evidence aggregation. We also observed a mild temporal asymmetry: phoneme cross-attention often peaks slightly before the envelope peak, whereas word cross-attention can lag behind it. Decoder self-attention mirrors these roles, with the phoneme decoder remaining highly local and the BART decoder exhibiting a broader triangular pattern consistent with longer-context conditioning.

These observations are qualitative and should not be interpreted as a definitive mechanistic account. Rather, they provide a plausible picture of how the model often segments time and allocates evidence during decoding. Although the analysis is qualitative, manual inspection of all 880 validation trials indicated that similar temporal chunking patterns recur throughout the validation set, suggesting that the example in Fig.~\ref{fig:attention} is representative rather than exceptional. Additional examples are provided in Appx.~\ref{attention extraction}: Fig.~\ref{fig:attention_triplet} shows encoder-decoder chunking and segment reuse across 20 randomly sampled trials, and Fig.~\ref{fig:attention_self} shows related encoder-derived chunking patterns across 4 multitask variants and 15 random trials.

\subsection{Word-level decoding tradeoffs: rescoring and inference speed}\label{subsec:inference_time}

Although the direct seq2seq model already provided the strongest phoneme-level performance, word-level output benefited substantially from an additional rescoring stage. As shown in Table~\ref{tab:decoding_results}, candidate generation and rescoring reduced WER from 25.6 $\pm$ 0.2\% to 19.4 $\pm$ 0.3\%. The strongest gains came from combining phoneme-consistency signals with linguistic plausibility, whereas language-model scoring alone was insufficient (Appx.~\ref{generation appendix}). This suggests that the candidate set contains substantially better hypotheses than those selected by greedy decoding, but that effective selection depends on preserving closeness to the neural evidence rather than optimizing language-model preference alone.

\begin{table}[h]
\centering
\footnotesize
\caption{Inference time on the test set (880 sentences; 5,166 words). Times are wall-clock.}
\label{tab:inference_time}
\tablebox{
\begin{tabular}{lccccc}
\toprule
\textbf{Model} & \textbf{HW} & \textbf{Sent/s} & \textbf{Words/s} & \textbf{ms/sent} & \textbf{ms/word} \\
\midrule
Transformer + MFCC + BART & H100 & 19.13 & 112.30 & 52.3 & 8.9 \\
RNN--CTC + MFCC + BART & H100 & 30.34 & 178.14 & 33.0 & 5.6 \\
RNN--CTC + MFCC + Willett LM & A100 + CPU & 1.57 & 9.20 & 638.6 & 108.7 \\
Transformer + rescoring pipeline & H100 & 3.12 & 18.32 & 320.4 & 54.6 \\
\bottomrule
\end{tabular}
}
\end{table}

We benchmarked wall-clock decoding on the full test split (880 trials; 5,166 words total). On a single H100 GPU, the direct Transformer + MFCC + BART model decoded at 19.1 sentences/s (112.3 words/s), completing the test set in 46\,s. Adding candidate generation and rescoring increased runtime to 282\,s, corresponding to 3.1 sentences/s. Even so, this configuration remained substantially faster than a Willett-style hybrid pipeline with WFST search and language-model rescoring, which was the slowest overall at 562\,s. These differences reflect the practical overhead introduced by multi-stage search and rescoring.

These measurements clarify the operating points of the different decoding strategies. Hybrid and heavily rescored pipelines achieve lower WER, but at substantially higher computational cost. In contrast, direct end-to-end seq2seq decoding offers much higher throughput and simpler deployment, while retaining strong phoneme-level fidelity and competitive word-level performance.

\subsection{Scaling with dataset size}\label{subsec:scaling}

\begin{figure}[h]
    \centering
    \includegraphics[width=0.9\linewidth]{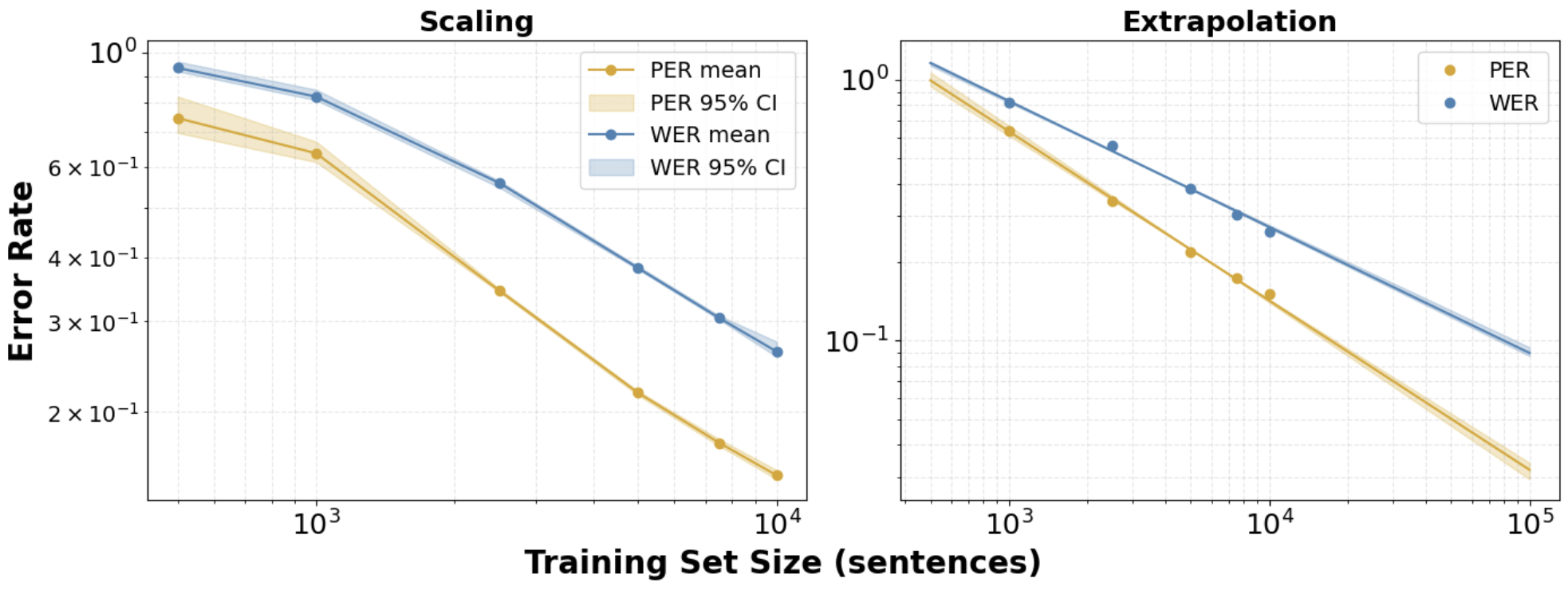}
    \caption{\textbf{Scaling behavior and power-law extrapolation.} \textbf{Left:} Phoneme (PER) and word (WER) error rates for six dataset fractions $\{0.05,\,0.10,\,0.25,\,0.50,\,0.75,\,1.00\}$, plotted as corresponding training-set sizes, with 95\% bootstrap confidence intervals across seeds. \textbf{Right:} Power-law models fitted to the multi-seed mean curve (excluding the 0.05 fraction) with bootstrap 95\% confidence bands and extrapolated performance up to 100,000 training trials.}
    \label{fig:scaling}
\end{figure}

Finally, we examined whether classical scaling-law behavior also appears in intracortical speech decoding. Using day-stratified subsampling of the 8,800 training trials at fractions $\{0.05, 0.10, 0.25, 0.50, 0.75, 1.00\}$, we retrained the best-performing seq2seq variant (Transformer + MFCC + BART, without rescoring) with 5 seeds per fraction while holding other training factors fixed. Figure~\ref{fig:scaling} shows that both PER and WER follow an approximately linear trend in log--log space, consistent with scaling behavior reported in other machine learning domains~\citep{scalingHestness,scalingKaplan}.

We fitted a power law of the form $e(N)=aN^b$ to the multi-seed mean curve and used bootstrap resampling to estimate parameter uncertainty. Because the 0.05 fraction appears to lie in a pre-asymptotic regime, the extrapolation shown in Figure~\ref{fig:scaling} excludes this point. Under that fit, increasing the dataset from approximately 10k to 20k sentences is projected to reduce PER from 14.3\% to 8.4\% and WER from 25.6\% to 18.3\%. A 100k-sentence dataset is further projected to reach approximately 2.9\% PER and 8.4\% WER.

These extrapolations should be interpreted cautiously. Power-law fits assume that the data distribution remains sufficiently stable as dataset size grows, whereas chronic intracortical recordings are subject to electrode drift, recalibration, and other nonstationarities. We therefore view these projections as illustrative estimates under relatively stable recording conditions rather than as guaranteed practical outcomes.

\section{Discussion}\label{sec:discussion}

Our study demonstrates that fully contextual seq2seq decoding is valuable not only as a speech decoder, but also as an analysis framework. In the present setting, it improves phoneme-level readout from intracortical activity, reveals a substantial dependence on session-specific calibration, and makes it possible to qualitatively ``decode the decoder'' through attention-based analysis of how neural evidence is segmented and reused over time.

A central takeaway of these results is that phoneme error rate (PER) and word error rate (WER) capture different aspects of system performance and should be interpreted as complementary rather than interchangeable metrics. WER remains the standard measure of communication performance, but it reflects both the neural decoder's accuracy and the strength of downstream linguistic correction. PER provides a more direct view of the neural-to-speech mapping itself. This distinction is particularly relevant for attempted-speech decoding from ventral premotor cortex, where the dominant neural information is expected to reflect articulatory and motor-planning variables more directly than high-level semantic content~\citep{Bouchard2013}.

Viewed through this lens, the main strength of our model is improved sublexical fidelity. Recent advancements, such as the DCoND system of Li et al. \cite {Li2025}, demonstrate that incorporating local phonetic context can improve CTC decoding: DCoND replaces monophone targets with context-dependent diphoneme-like units and explicitly models transitions, motivated by evidence that phoneme representations in neural space depend on surrounding context, especially the preceding phoneme. Our results push this principle further. By conditioning each output on the full neural input sequence and the generated history, the seq2seq model captures broader temporal dependencies and cross-level interactions between phoneme and word prediction, yielding a lower PER than both monophone-based and diphoneme-enriched CTC pipelines.

At the same time, the strongest reported WERs are obtained by heavier multi-stage systems that apply strong linguistic refinement after neural decoding. Such refinement can greatly improve transcription accuracy, but it can also obscure whether gains arise from better neural readout or from stronger post hoc language correction. For clinical deployment, it is therefore valuable to preserve decoding pipelines that remain transparently anchored to the neural evidence while still leveraging linguistic structure where appropriate, particularly for out-of-distribution phrases where excessive correction could distort user intent.

The architectural ablations further suggest that representation shaping matters. Auxiliary MFCC supervision improved PER across both recurrent and seq2seq model families, indicating that even coarse acoustic structure can regularize early neural representations. At the same time, our seq2seq model achieved state-of-the-art PER even without MFCC supervision, indicating that the contextual decoding formulation itself is the primary driver of the gain. Adding the word-level decoder further improved phoneme decoding in the seq2seq family, suggesting that higher-level linguistic supervision can beneficially shape sublexical representations. Additional ablations further suggest that pretrained BART initialization provides only modest benefit in this regime, implying that much of the useful structure is learned from the supervised neural-decoding task itself rather than imported directly from large-scale language pretraining.

Our results also indicate that session variability is an important practical factor in this decoding setting. NHS consistently outperformed both the linear day transform introduced by Willett et al. \cite{Willett2023} and models without any day-specific transform, indicating that effective calibration can require more flexibility than global alignment alone. The held-out-day experiments further showed that reusing a nearby-day transform allows nontrivial decoding on unseen days, but that the penalty grows substantially with temporal distance. Thus, NHS reduces but does not eliminate the calibration burden. In its current formulation, the transform also assumes that day identity is known at inference time, which limits direct deployment realism. 

The attention analyses provide a complementary perspective on how the model solves the decoding problem. Across the validation set, the encoder repeatedly developed temporally localized chunking patterns, while phoneme and word decoders re-used these segments differently. The phoneme decoder showed more localized evidence retrieval and self-attention, whereas the word decoder aggregated more broadly over segments and history. These patterns should not be treated as a definitive mechanistic account of speech coding in cortex, but they do suggest a structured hierarchy of evidence aggregation and provide a qualitative window into how the model organizes neural evidence over time.

This study has limitations. All experiments were performed on a single participant and a single cortical region, and decoding was evaluated offline rather than in a streaming setting. The day-specific calibration module assumed known session identity at inference time, and the attention analyses remained qualitative even though the chunking pattern was observed broadly across the validation set. Finally, although the proposed model achieved state-of-the-art PER, its best WER remained above that of the strongest hybrid and LLM-assisted pipelines.

These results suggest several directions for future work. First, improved phoneme-level readout motivates tighter integration with downstream word decoding, for example via structured phoneme-to-word search (WFST-style) or other constrained decoding schemes that preserve closeness to neural evidence. Second, robustness across days remains a central challenge: replacing explicit day identity with inferred session embeddings, calibration from limited adaptation data, or continuous/test-time adaptation could better track drift without manual session labels. Third, the recurrent chunking observed in attention patterns motivates streaming or limited-lookahead variants of seq2seq decoding that trade controlled latency for much of the offline performance.

In parallel, practical deployment raises ethical and safety considerations. Decoding errors can misrepresent user intent, particularly for out-of-distribution phrases such as names or novel strings, motivating interfaces that expose calibrated uncertainty, enable rapid user correction, and provide conservative fallbacks. Because neural signals are highly sensitive, strong protections are needed to ensure decoding occurs only with explicit consent and intentional engagement, including reliable ``off'' states and clear user control over when decoding is active. From a privacy perspective, reducing reliance on cloud-based computation is desirable; lightweight end-to-end decoding pipelines provide a practical foundation for on-device, privacy-preserving operation, even if higher-accuracy multi-stage refinement remains useful in some offline settings.

Overall, our results show that contextual seq2seq decoding improves phoneme-level intracortical speech readout, while highlighting session variability as a major remaining barrier and revealing recurrent temporal organization that can inform future decoder design.


\bibliographystyle{ieeetr}

\bibliography{bib}

\appendix
\clearpage

\section{Statements}
\section*{Ethics Statement}
This study uses the publicly available intracortical dataset of Willett et al.\cite{Willett2023Dataset}; no new human data were collected. The original dataset was gathered under IRB oversight with informed consent, and all analyses were performed on de-identified recordings. We complied with the dataset license and did not attempt re-identification or linkage with external sources. Our models are intended to advance scientific understanding of neural speech decoding and are not suitable for clinical, legal, or surveillance use. Because attempted/inner speech may raise privacy concerns, we emphasize safeguards discussed in Section~\ref{sec:discussion}: intention-based operation (e.g., “go” signals, robust ``off'' states), calibrated uncertainty and user correction, conservative deployment practices, and opt-in/opt-out data handling. We disclose potential risks of misuse (misinterpretation of intent, unauthorized decoding) and recommend that any future deployment include explicit consent, human oversight, and protections against non-consensual use.

\section*{Reproducibility Statement}
We took multiple steps to ensure reproducibility. The dataset source and all preprocessing are documented in Section~\ref{sec:dataset}. The full model architecture, the NHS day transform, training objectives, training protocol, and evaluation metrics are specified in Section~\ref{sec:model} and Section~\ref{sec:training}. Hyperparameters are listed in Appendix~\ref{model_parameters}. The scaling-law protocol (fractions and fitting) appears in Section~\ref{subsec:scaling}. Attention extraction procedures are described in Appendix~\ref{attention extraction}. We report means and standard deviations over five seeds in Table~\ref{tab:decoding_results} and provide exact splits (stratified by day) in our code release. 

\section{Qualitative error analysis}\label{subsec:error}

\begin{table}[h]
\centering
\footnotesize
\caption{Example word-level predictions of Transformer+MFCC+BART model (no rescoring) from 10th, 50th (median) and 90th WER percentiles.}
\label{tab:predictions_wer}
\tablebox{
\begin{tabular}{rrll}
\toprule
\textbf{WER} & \textbf{\%ile} & \textbf{Target sentence} & \textbf{Predicted text} \\
\midrule
0.0 & 0 & Do you know where it might have gone? & Do you know where it might have gone. \\
0.0 & 0 & I am an artist, lost in my own vision. & I am an artist, lost in my own vision. \\
0.0 & 0 & Read the decision below. & Read the decision below. \\
\midrule
0.2 & 50 & I don't think so anymore. & I don't think so many. \\
0.2 & 50 & Just \textbf{way} in the back. & Just \textbf{why} in the back? \\
0.2 & 50 & Sometimes they're not very \textbf{open}. & Sometimes they're not very \textbf{hard}. \\
\midrule
0.6 & 90 & \textbf{Fifty} nine \textbf{kilometers} \textbf{per} gallon. & \textbf{Fifteen} nine \textbf{automobiles} \textbf{for} gallon. \\
0.6 & 90 & We were \textbf{promised} \textbf{civil liberties}. & We were \textbf{most} \textbf{several places}. \\
0.6 & 90 & \textbf{Special rules} for employment \textbf{cases}. & \textbf{Paces} for employment \textbf{taxes}. \\
\bottomrule
\end{tabular}
}
\end{table}

Table~\ref{tab:predictions_wer} shows representative predictions of Transformer + MFCC + BART model (no rescoring) at the 10th, 50th (median), and 90th WER percentiles. Errors at the median and tail are dominated by phonetic confusions with high acoustic similarity (e.g., \emph{way}$\rightarrow$\emph{why}, \emph{fifty}$\rightarrow$\emph{fifteen}; \emph{cases}$\rightarrow$\emph{taxes}), often preserving sentence rhythm but shifting semantics. This suggests that the neural signal supplies strong sub-lexical acoustic evidence while semantic disambiguation remains limited without explicit rescoring.

\section{BART initialization and training: Linguistic prior impact}
\label{bart init}

We examined whether word-level linguistic priors encoded in the pretrained BART decoder influence decoding performance. In Stage~2, we compared three configurations: (i) initializing the BART decoder from the \texttt{bart-base} checkpoint while freezing its first three layers to preserve pretrained linguistic structure, (ii) initializing from \texttt{bart-base} and finetuning all layers, and (iii) replacing the decoder with an identical BART architecture initialized with random weights. In all conditions, the same pretrained BART tokenizer was used, so the subword vocabulary and segmentation were held fixed.

As shown in Table~\ref{tab:bart_init}, all strategies achieve highly similar phoneme and word error rates. The fully finetuned pretrained model yields the best WER, but the difference relative to random initialization is small. Notably, the randomly initialized decoder performs on par with the pretrained variants, despite lacking any large-corpus language modeling priors in its weights. This suggests that the linguistic structure exploited during decoding arises primarily from patterns in the supervised dataset and from the encoder’s neural representations, rather than from the pretrained BART language model itself. 

Comparison to Table~\ref{tab:decoding_results} shows that adding the word-level BART decoder—regardless of how its weights are initialized—reduces phoneme error rates relative to a phoneme-only baseline. This indicates that the encoder representations are shaped by word-level supervision even when the decoder itself carries no pretrained linguistic knowledge. Taken together, these results imply that our transformer encoder can learn higher-level, linguistically structured representations directly from neural activity and task data, with pretrained BART weights providing at most a modest additional benefit in this regime.

\begin{table}[h]
\centering
\footnotesize
\caption{Performance comparison between different strategies of BART initialization and training. Results  reported as mean ± SD across 5 global seeds}
\label{tab:bart_init}
\tablebox{
\begin{tabular}{lcc}
\toprule
\textbf{Model} & \textbf{PER (\%)} & \textbf{WER (\%)} \\
\midrule
Transformer Seq2Seq + MFCC + BART &  &  \\
\quad BART pretrained + Freeze first 3 layers & 14.3 $\pm$ 0.3 & 25.6 $\pm$ 0.2 \\
\quad BART pretrained + No freezing & 14.6 $\pm$ 0.4 & \textbf{25.2 $\pm$ 0.1} \\
\quad BART random init + No freezing & \textbf{14.2 $\pm$ 0.2} & 25.8 $\pm$ 0.4  \\

\bottomrule
\end{tabular}
}
\end{table}

\section{Generation Strategies and Scoring Analysis}
\label{generation appendix}

We employed two complementary decoding configurations to construct a rich hypothesis space for the rescoring stage:

\begin{itemize}
    \item \textbf{Beam-20 (deterministic).}  
    A standard beam search with \texttt{num\_beams}=20, producing 20 high-likelihood candidates per trial. This method explores the most probable regions of the model’s output distribution in a deterministic manner.

    \item \textbf{Nucleus-128 (stochastic).}  
    A high-diversity sampling setup that generates 128 candidates using nucleus sampling (\texttt{top\_p}=0.95, \texttt{temperature}=1.0, \texttt{top\_k}=0). Unlike beam search, this strategy introduces stochasticity and explores a broader portion of the distribution, often producing phrasings that would not appear under purely probability-maximizing generation.
\end{itemize}

\begin{table}[h]
\centering
\footnotesize
\caption{Word-level performance comparison between strategies of candidate selection. Score-based results are obtained by selection of candidates generated using beam search and stochastic sampling strategies. All results reported as mean ± SD across 5 global seeds}
\label{tab:gen_score}
\tablebox{
\begin{tabular}{lcc}
\toprule
\textbf{Model} & \textbf{WER (\%)} \\
\midrule
Transformer Seq2Seq + MFCC + BART &  &  \\
\midrule
\quad Greedy decoding  & 25.6 $\pm$ 0.2 \\
\quad Beam search (top result) & 24.8 $\pm$ 0.4 \\
\midrule
\quad PER score only & 23.2 $\pm$ 0.4  \\
\quad Phoneme head score only & 22.6 $\pm$ 0.2  \\
\quad LLM score only & 27.2 $\pm$ 0.4  \\
\quad All scores combined & 19.4 $\pm$ 0.3  \\
\quad Oracle & 14.5 $\pm$ 0.2  \\

\bottomrule
\end{tabular}
}
\end{table}

Table~\ref{tab:gen_score} summarizes the effect of different candidate selection approaches. The default greedy decoding provides a strong baseline, while beam search offers only a modest improvement when simply the highest overall-probability beam output is chosen. More substantial gains arise when candidates are rescored using phoneme-level evidence (PER-score and phoneme-score), especially compared to LLM-only scoring--highlighting the weakness of BART decoder to differentiate in the acoustic space as shown in \ref{subsec:error}. The best overall performance is obtained when all scoring components are combined with ratio PhonemeHead:PER:LLM 9:4:5, lowering the final WER from 25.6\% to 19.4\% PER (delta 6.2 pp.). The word-level performance could theoretically be improved further as the oracle (always choose the best) WER is 14.5\%.

\section{Attention extraction and further examples}
\label{attention extraction}

\begin{figure}[h]
    \centering
    \includegraphics[width=1.0\linewidth]{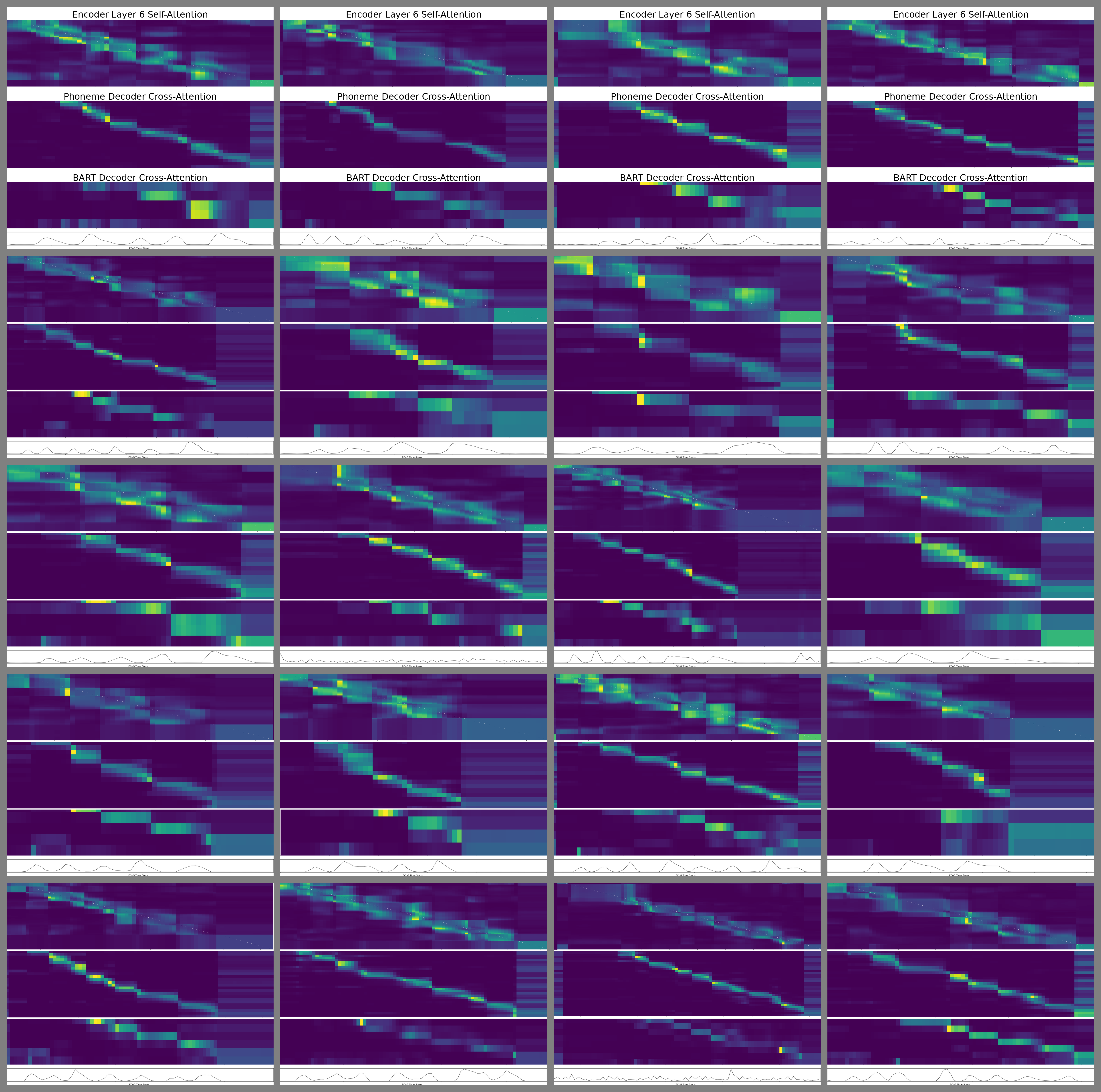}
    \caption{\textbf{Self and cross attention: Usage of temporal chunking.} Each cell shows vertically stacked 1) Encoder self-attention in the last layer, 2) Phoneme Decoder cross-attention, 3) BART word decoder cross-attention, 4) Audio envelope, obtained from Transformer Seq2Seq + MFCC + BART model, visualized for 20 randomly chosen validation trials. The pattern of temporal chunking of neural signal in the encoder and the downstream usage of those chunks by both decoders is prevalent across all samples - as well as the temporal lag of word decoder relative to the phoneme decoder, indicating evidence accumulation.}
    \label{fig:attention_triplet}
\end{figure}

\begin{figure}[h]
    \centering
    \includegraphics[width=1.0\linewidth]{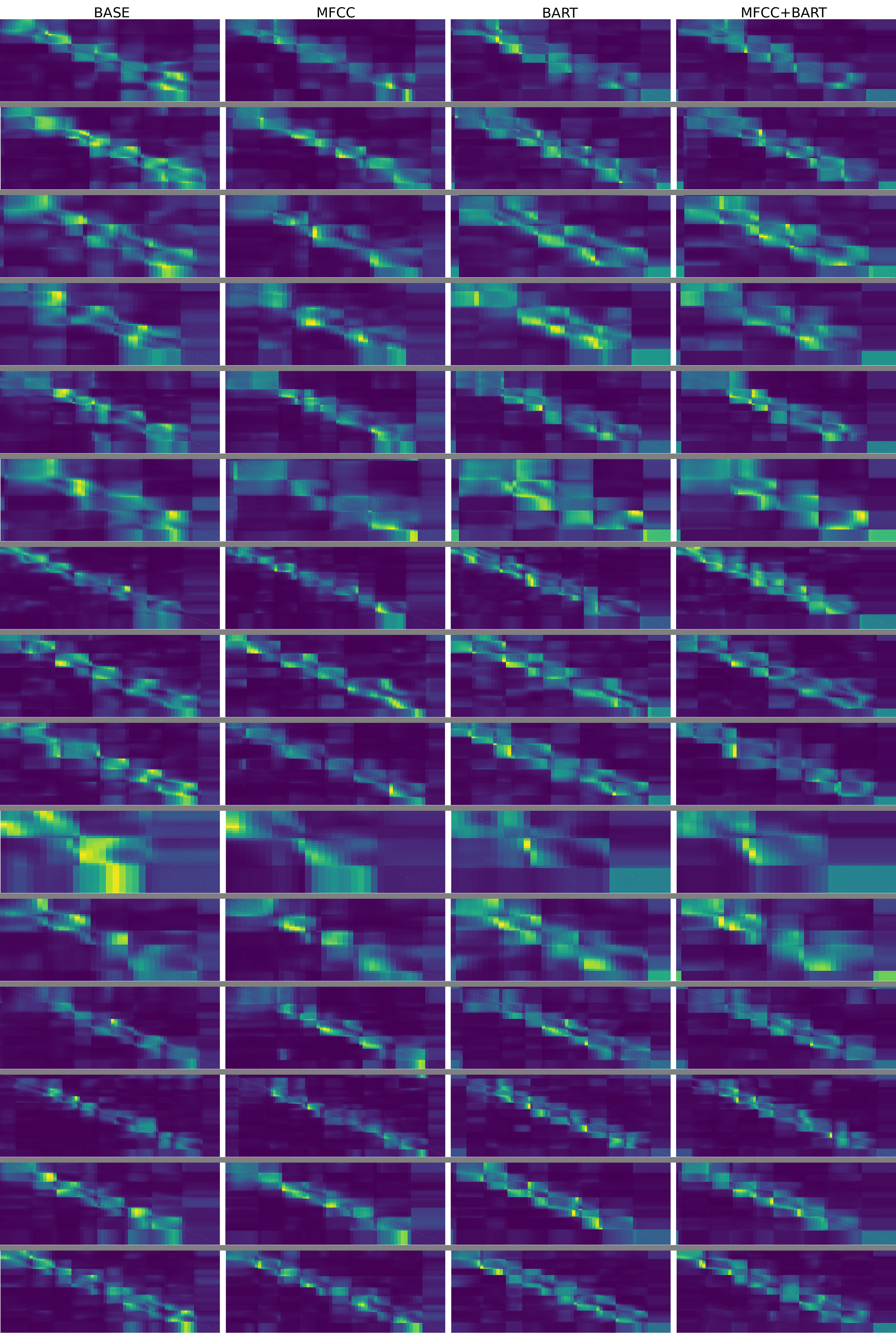}
    \caption{\textbf{Self-attention: Temporal chunking.} Self-attention map of the last (6th) layer of encoder across model variants (columns) visualized for random 15 trials (rows). The representation pattern of temporal chunking of the neural signal into "boxes" is prevalent across the trials regardless of the auxiliary objectives}
    \label{fig:attention_self}
\end{figure}

To analyze model behavior, we extracted attention maps from the encoder and both decoders. The PyTorch TransformerEncoder and phoneme TransformerDecoder were modified setting nn.MultiheadAttention class attribute \texttt{return\_weights=True} to return attention weights for each layer. For the word decoder, we used the Hugging Face BART implementation with \texttt{output\_attentions=True} to visualize attention heads across tokens. These attention maps allow us to interpret which neural timesteps contribute to specific phoneme or word predictions, shedding light on model alignment and representation dynamics. Figures \ref{fig:attention_triplet} and \ref{fig:attention_self} provide further examples of the observed patterns among randomly sampled 20 and 15 validation trials respectively.

\section{Transformer model parameters}
\label{model_parameters}

\begin{table}[h]
\centering
\footnotesize
\caption{Training and model parameters for the two-stage training process of Transformer+MFCC+BART model. Stage 1 focuses on pre-training the ECoG encoder with an auxiliary MFCC prediction task alongside the main phoneme decoding task. Stage 2 introduces a BART-based word decoder head and fine-tunes the entire model jointly on phoneme and text decoding. The Stage 2 column only lists values in Multi-Task Setup section, as only then they differ from Stage 1.}
\label{tab:training_params}
\resizebox{\textwidth}{!}{%
\tablebox{
\begin{tabular}{@{}lll@{}}
\toprule
\textbf{Parameter} & \textbf{Stage 1 Value} & \textbf{Stage 2 Value} \\
\midrule
GPU & Nvidia H100 80GB of HBM3 memory & \\
\midrule
\multicolumn{3}{c}{\textbf{Core Transformer Architecture}} \\
\midrule
Model dimension (\texttt{d\_model}) & 512 & \\
Attention heads (\texttt{n\_head}) & 8 & \\
Encoder layers & 6 & \\
Decoder layers & 6 & \\
Feed-forward dimension & 2048 & \\
Activation function & GELU & \\
\midrule
\multicolumn{3}{c}{\textbf{Input Processing \& Feature Extraction}} \\
\midrule
ECoG features & 256 (128 channels x 2 features per-channel) & \\
Feature extractor & Binned Attention Conv & \\
Conv kernel size & 5 & \\
Downsampling strategy & Convolutional & \\
Downsampling factor & 4 & \\
Day adaptation & NHS & \\
Number of days & 24 (all) & \\
\midrule
\multicolumn{3}{c}{\textbf{Regularization \& Augmentation}} \\
\midrule
Dropout & 0.4 & \\
Time masking probability & 0.3 & \\
Max time mask length & 25 steps & \\
Max time mask proportion & 20\% & \\
Channel masking probability & 0.25 & \\
Max channels masked & 15 electrodes & \\
\midrule
\multicolumn{3}{c}{\textbf{Multi-Task Setup}} \\
\midrule
Training objective & Joint phoneme decoding \& MFCC prediction & Joint phoneme \& text decoding \\
Phoneme decoder loss weight & 1.0 & 1.0 \\
MFCC auxiliary task & Enabled & Disabled \\
MFCC head type & Linear & N/A \\
MFCC aux. loss weight & 0.001 & N/A \\
Aux. head input layer index & 1 & N/A \\
BART text decoder task & Disabled & Enabled \\
BART model type & N/A & \texttt{bart-base} \\
BART loss weight & N/A & 1.0 \\
BART freezing strategy & N/A & Freeze first 3 decoder layers \\
\midrule
\multicolumn{3}{c}{\textbf{Optimizer \& Scheduler}} \\
\midrule
Optimizer & AdamW & \\
Learning rate & 1e-4 & \\
Weight decay & 1e-3 & \\
Scheduler & ReduceLROnPlateau & \\
Scheduler metric & \texttt{val\_per\_sg\_epoch} & \\
Scheduler factor & 0.5 & \\
Scheduler patience & 10 epochs & \\
\midrule
\multicolumn{3}{c}{\textbf{Training Details}} \\
\midrule
Batch size & 4 & \\
Gradient accumulation & 5 steps & \\
Effective batch size & 20 & \\
Epochs & 200 & \\
Precision & 32-bit float & \\
ECoG Encoder frozen & No & \\
Phoneme Decoder frozen & No & \\
\bottomrule
\end{tabular}%
}
}
\end{table}

\end{document}